\documentclass[10pt,twocolumn,letterpaper]{article}

\usepackage[pagenumbers]{cvpr} 

\usepackage{amsmath}
\usepackage{makecell}
\usepackage{graphicx}
\usepackage{multirow}
\usepackage{tabu}
\usepackage{pifont}
\usepackage{adjustbox}
\usepackage{float}

\newcommand{\model}{GenMOJO}
\newcommand{\dataset}{MOSE-PTS}

\definecolor{cvprblue}{rgb}{0.21,0.49,0.74}
\usepackage[pagebackref,breaklinks,colorlinks,allcolors=cvprblue]{hyperref}

\def\paperID{1513} 
\def\confName{CVPR}
\def\confYear{2025}

\title{Generative 4D Scene Gaussian Splatting with Object View-Synthesis Priors}

\author{Wen-Hsuan Chu$^{1*}$\hspace{0.3cm}Lei Ke$^{1*}$\hspace{0.3cm}Jianmeng Liu$^{1*}$\hspace{0.3cm}Mingxiao Huo$^{1}$\hspace{0.3cm} \\ Pavel Tokmakov$^{2}$\hspace{0.3cm}Katerina Fragkiadaki$^{1}$ \\ \\
\textsuperscript{1}Carnegie Mellon University\hspace{1.0cm} \textsuperscript{2}Toyota Research Institute
}

\begin{document}
\maketitle
\let\thefootnote\relax\footnotetext{*Equal contributions}
\begin{abstract}

We tackle the challenge of generating dynamic 4D scenes from monocular, multi-object videos with heavy occlusions, and introduce \model{}, a novel approach that integrates rendering-based deformable 3D Gaussian optimization with generative priors for view synthesis.  While existing models perform well on novel view synthesis for isolated objects, they struggle to generalize to complex, cluttered scenes. 
To address this, \model{} decomposes the scene into individual objects, optimizing a differentiable set of deformable Gaussians per object. This object-wise decomposition  allows leveraging object-centric diffusion models to infer unobserved regions in novel viewpoints. It performs joint Gaussian splatting to render the full scene, capturing cross-object occlusions,  and enabling occlusion-aware supervision.  
To bridge the gap between object-centric priors and the global frame-centric coordinate system of videos, \model{} uses differentiable transformations that align generative and rendering constraints within a unified framework. The resulting model generates 4D object reconstructions over space and time, and produces accurate 2D and 3D point tracks from monocular input.
Quantitative evaluations and perceptual human studies confirm that \model{} generates more realistic novel views of scenes and produces more accurate point tracks compared to existing approaches. Project page: \url{https://genmojo.github.io/}.
\end{abstract}    

\section{Introduction}
\label{sec:intro}

Humans effortlessly perceive complete, coherent objects and their motions from videos even though they only observe the pixels on the front surface of a dynamic scene, which constantly changes due to motion, occlusion, and reappearance. Recovering persistent and accurate 4D representations from monocular videos with multiple objects is a highly under-constrained problem. However, it is critical for fine-grained video understanding, visual imitation in robotics~\cite{heiden2022inferring}, and building causal world models of intuitive physics~\cite{xie2023physgaussian} that can simulate the consequences of actions.

Recent view-predictive generative models~\cite{zeronvs, liu2023zero} provide powerful priors for novel view synthesis. 
However, existing video-to-4D generation works~\cite{dreamscene4d,jiang2023consistent4d,yin20234dgen,zeng2024stag4d}~based on these generative models often struggle to maintain temporal coherence and accurate 3D geometry in complex multi-object scenes due to the intricate dynamics involved, such as heavy occlusions and fast motion.

\begin{figure*}[!t]
    \centering
    \includegraphics[width=0.98\linewidth]{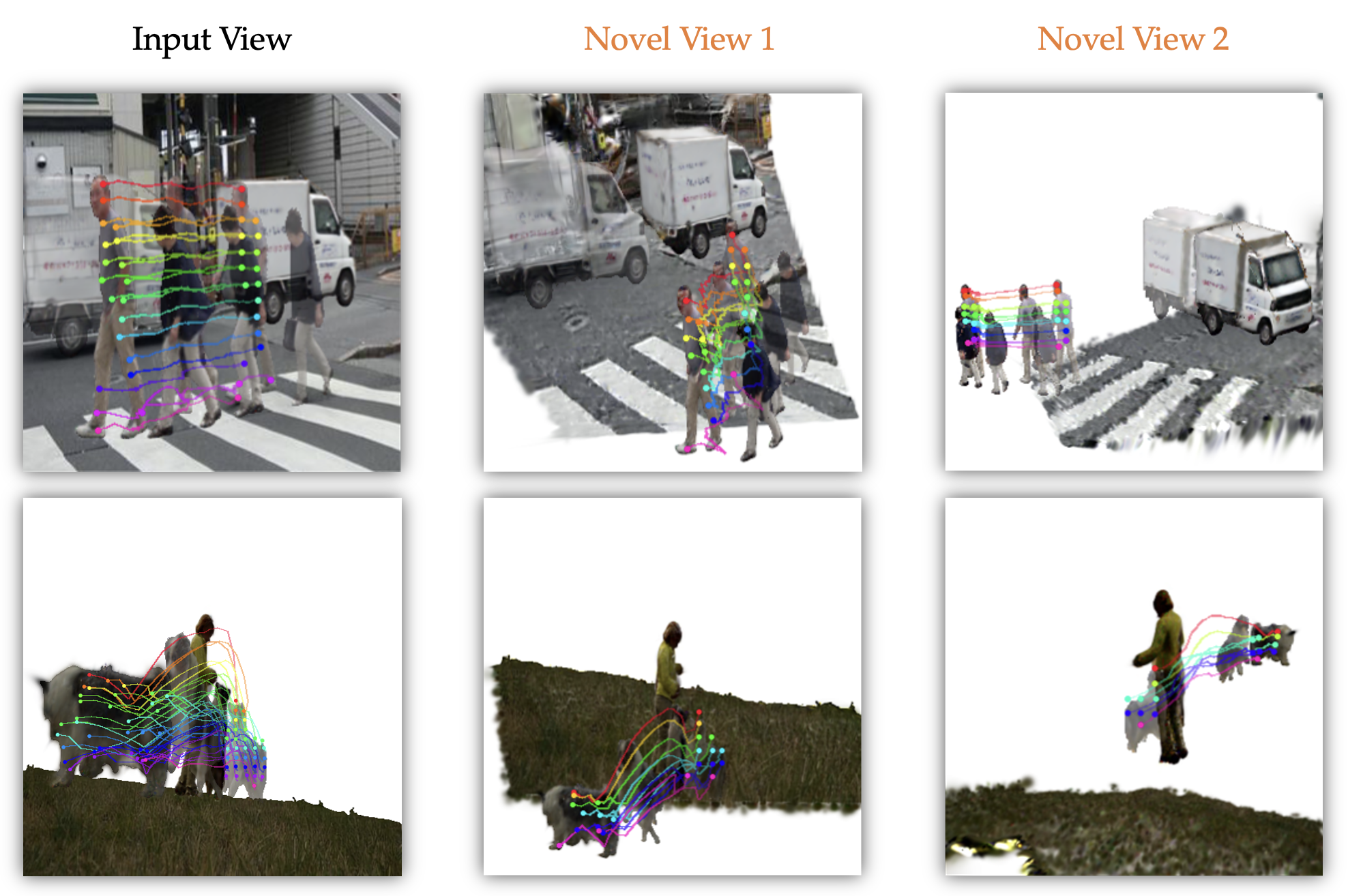}
    \caption{ \textbf{\model{} addresses video-to-4D generation for real-world scenes with many moving objects and heavy occlusions.} It creates complete 4D scene reconstructions, supports $360^\circ$ novel view synthesis, and accurately tracks how points move over time. We show examples of rendered images from different viewpoints and time steps, highlighting the motion of a single object for clarity. }
    \label{fig:in-the-wild-teaser}
\end{figure*}

In this paper, we propose \model{}, a compositional method that enables \textbf{M}ulti-\textbf{O}bject \textbf{JO}int splatting for 4D scene \textbf{GEN}eration from monocular videos by combining frame-centric test-time rendering error optimization with object-centric generative priors. \model{} decomposes a scene into individual foreground objects and background, representing each object with a differentiable and deformable set of 3D Gaussians. We optimize the parameters of these Gaussians by: 
\begin{enumerate}
\item \textit{Jointly} splatting all Gaussians to compute rendering errors in the input video frames, accounting for occlusions; 
\item \textit{Individually} rendering each object’s Gaussians from unobserved viewpoints to optimize a score distillation (SDS) objective. 
\end{enumerate}
To bridge frame-centric and object-centric coordinate frames, \model{} infers \textbf{differentiable transformations} that align the object-centric Gaussians with the frame-centric Gaussians. 
Unlike previous methods that focused on single-object~\cite{ren2023dreamgaussian4d,jiang2023consistent4d} or simple multi-object videos~\cite{dreamscene4d},~\model{} is designed to handle scenes with heavy occlusions, crowded environments, and intricate dynamics, as those shown in Figure~\ref{fig:in-the-wild-teaser}. 

We evaluate \model{} on multi-object videos from DAVIS and a subset of MOSE~\cite{ding2023mose}, for which we manually annotated accurate point tracks. 
We compare against state-of-the-art NeRF and Gaussian-based 4D generation methods~\cite{jiang2023consistent4d,dreamscene4d,wang2024shape}. Our results show substantial improvements in both 4D rendering quality and motion accuracy. In addition, we conducted human evaluations to assess the perceptual realism of our results, which further support our quantitative findings.
\section{Related work}
\label{s:related}

\paragraph{Dynamic Scene Reconstruction.}
Dynamic scene reconstruction aims to lift the visible parts of a video to a dynamic 3D representation for novel view synthesis. These methods often take videos where a large number of camera views exist as input~\cite{luiten2023dynamic,wu20234d,yang2023deformable3dgs}, alleviating the issue of partial observability. As a result, many of these methods can not be readily applied to monocular videos, as no constraints exist in unobserved viewpoints. These methods fall into two large categories. Dynamic Neural Radiance Field (NeRF) based methods~\cite{pumarola2021d,park2021nerfies,li2022neural,liu2023robust,busching2023flowibr} extend NeRFs~\cite{mildenhall2020nerf} to dynamic scenes and model a dynamic scene using grid-based, voxel-based representations~\cite{lombardi2019neural, cao2023hexplane, liu2020neural} or through deformation networks~\cite{cao2023hexplane,fridovich2023k}. Recently, Gaussian Splatting ~\cite{kerbl20233d,luiten2023dynamic,wu20234d} has gained great attention in the field of neural rendering. Compared to NeRF-based methods, Gaussian Splatting represents the scene as explicit Gaussians and uses efficient rasterization techniques for rendering, greatly improving the computation efficiency. There have also been efforts to model the visible parts of a video scene from monocular inputs, but they do not attempt to model the unobserved parts~\cite{wang2024shape,lei2024mosca,stearns2024dynamic,wang2024gflow}. Compared to these works, we adopt dynamic Gaussians to synthesize both the visible and unobserved parts of the video scene. 

\paragraph{Dynamic Scene Generation.}
In contrast to dynamic scene reconstruction, dynamic scene generation is concerned with modeling a complete video scene across both visible and unobserved viewpoints, often operating under monocular or sparse view settings. Existing text-to-4D~\cite{singer2023text,bahmani20234d,ling2023align,bahmani2024tc4d} or image-to-4D~\cite{yang2024beyond,zheng2023unified,ren2023dreamgaussian4d} works use diffusion models~\cite{liu2023zero} to condition on text or image inputs and generate synthetic videos, then use score distillation sampling (SDS)~\cite{poole2022dreamfusion} losses to supply constraints in unobserved viewpoints to synthesize complete 4D representations using dynamic NeRFs or Gaussians. Many existing video-to-4D works simplify this problem by assuming that the input consists of a single object undergoing small movements~\cite{ren2023dreamgaussian4d,jiang2023consistent4d,gao2024gaussianflow,pan2024fast,yin20234dgen,zeng2024stag4d}. More recently, this line of work has been extended to multi-object video by first decomposing the video into multiple object tracks, optimizing them independently, and then using depth models~\cite{depthanything} to recompose the optimized 4D objects to re-form the original video scene~\cite{dreamscene4d}. 

The work most closely related to ours is DreamScene4D~\cite{dreamscene4d}, which also leverages object-centric generative priors to produce 4D scene completions from monocular video. However, their approach inpaints and optimizes the 3D deformable Gaussians for each object independently and only combine them during a final rendering step to infer object depth ordering. This strategy struggles with cross-object occlusions, as rendering objects separately fails to account for occluders. As a result, their method often produces unnatural, inter-penetrating artifacts in scenes with multiple heavily occluded objects, as demonstrated in our experiments. In contrast, our approach unifies image-based and object-centric constraints within a single gradient-based optimization framework. By jointly splatting all object Gaussians, we accurately model cross-object occlusions and interactions, while still imposing strong appearance priors by rendering each object independently.

\begin{figure*}[!t]
    \centering
\includegraphics[width=0.98\linewidth]{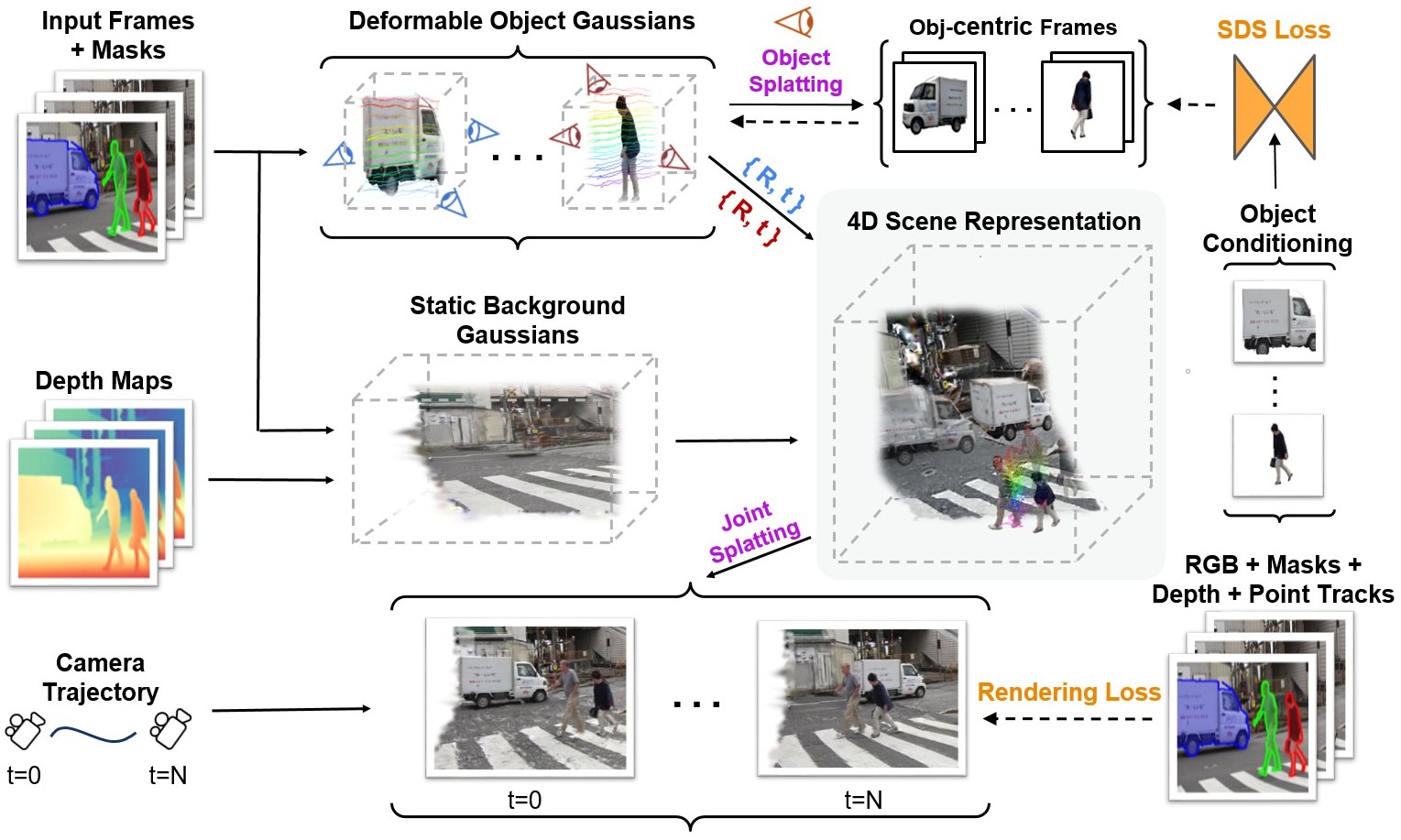}
    \caption{\textbf{Overview of \model{}.} Given a monocular video, its camera pose trajectory, and segmentation tracks for each dynamic object, our model generates a set of deforming 3D Gaussians per object. This is achieved through object-centric Score Distillation Sampling (SDS) and frame-centric joint Gaussian splatting, which accounts for cross-object occlusions by imposing rendering losses on rendered instance masks, point tracks, depth, and RGB appearance. Differentiable transformations are estimated for each object and each timestep to map the object-centric Gaussians into the camera coordinate frame.}
    \label{fig:method}
\end{figure*}

\paragraph{Point Tracking via 4D Reconstruction.}
Reconstructing 4D scenes from monocular videos enables point trajectory extraction through test-time optimization, where dynamic radiance fields are fit to observed video frames without requiring explicit point tracking annotations~\cite{wang2023tracking,luiten2023dynamic,wang2024shape,seidenschwarz2024dynomo,dreamscene4d}. In contrast, data-driven 2D and 3D point trackers trained on synthetic datasets often outperform these optimization-based methods on standard benchmarks~\cite{karaev2023cotracker}.

However, in our experiments, we find that the performance gap narrows significantly in multi-object scenes with heavy occlusions~\cite{ding2023mose}. Feedforward trackers struggle with severe occlusions, whereas our method maintains robust tracking by leveraging generative models to infer the \textit{complete} 4D object geometry, even in unobserved or occluded regions.
\section{4D Generation from Monocular Video}
\model{} represents 4D scenes using deformable 3D Gaussians~\cite{kerbl20233d,wu20234d}  and employs diffusion-based generative models~\cite{liu2023zero} to enable novel view synthesis. In Section~\ref{sec:background}, we provide background on deformable 3D Gaussians and their optimization using score distillation sampling. Our full model architecture is presented in Section~\ref{sec:model_overview}.

\subsection{Preliminaries: Generative Gaussian Splatting}
\label{sec:background}

3D Gaussian Splatting~\cite{kerbl20233d} represents a scene as a set of colored 3D Gaussians.
Each Gaussian is defined by its position \(\mathbf{p} = \{x, y, z\} \in \mathbb{R}^3\),  size \(\mathbf{s} \in \mathbb{R}^3\) (given as standard deviations), and orientation \(\mathbf{q} \in \mathbb{R}^4\) encoded as a quaternion. For efficient  $\alpha$-blending during rendering, each Gaussian also has an opacity value  \(\alpha \in \mathbb{R}\) and a color representation \(\mathbf{c}\) expressed in spherical harmonics (SH) coefficients. 

\paragraph{Generative 3D Gaussian Splatting through SDS.} 
Score Distillation Sampling (SDS)~\cite{poole2022dreamfusion} is widely used for optimizing 3D Gaussian representations in text-to-3D and image-to-3D generation by leveraging a diffusion prior to help infer plausible novel views. DreamGaussian \cite{tang2023dreamgaussian}, for instance, uses Zero-1-to-3 \cite{liu2023zero}, a model that predicts a novel view given a reference image and a relative camera pose, to lift 2D inputs into 3D Gaussian representations from a single frame. The optimization objective combines a differentiable rendering loss and an SDS loss:
\begin{equation}
    \nabla_\phi \mathcal{L}_\text{SDS}^\text{t}=\mathbb{E}_{t, \tau, \epsilon, p}\left[w(\tau)\left(\epsilon_\theta\left(\hat{I}^p_t ; \tau, I_1, p\right)-\epsilon\right) \frac{\partial \hat{I}^p_t}{\partial \phi}\right], \nonumber
    \label{eq:sds_loss_image}
\end{equation} 
where $t$ is the timestep indices, $w( \tau )$ is a weighting function for denoising timestep $\tau$, $\phi\left(\cdot\right)$ represents the Gaussian rendering function, $\hat{I}^p_t$ is the rendered image under camera pose $p$, $\epsilon_\theta \left( \cdot \right)$ is the predicted noise by the diffusion model, and $\epsilon$ is the target noise. 

\paragraph{Modeling Gaussian Deformations in Videos.}
To capture motion over time, DreamGaussian4D \cite{ren2023dreamgaussian4d} models the deformation of 3D Gaussians using learnable parameters for each frame: \textbf{(1)} a 3D displacement for each timestep $\mu_t = \left( \mu x_t, \mu y_t, \mu z_t \right)$, \textbf{(2)} a 3D rotation for each timestep, represented by a quaternion $\mathcal{R}_t = \left( qw_t, qx_t, qy_t, qz_t \right)$, and \textbf{(3)} a 3D scale for each timestep $s_t = \left( sx_t, sy_t, sz_t \right)$. The color (SH coefficients) and opacity remain fixed across time and are initialized from the first frame.  
To estimate these per-frame deformations, a deformation network $D_{\theta} ( G^{obj}_0, t )$ is trained for each object. 
Given the static Gaussians $G^{obj}_0$ from the first frame and a target timestep $t$, the network outputs the full 10-D deformation vector. The network is trained with both rendering and SDS losses over all video frames, ensuring temporally consistent 4D Gaussians.

\subsection{\model{}}
\label{sec:model_overview}

Figure~\ref{fig:method} illustrates the architecture and workflow of \model{}, a test-time 4D Gaussian splatting framework that generates complete dynamic scenes from monocular multi-object videos. \model{} jointly optimizes RGB, optical flow, depth, and trajectory-based rendering objectives by combining scene-centric and object-centric modeling.

Given a monocular video, we estimate camera poses and per-frame depth using recent video geometry methods~\cite{megasam, depthCrafter}, and segment and track objects using state-of-the-art tools~\cite{ren2024grounded, sam_hq, sam2}. To constrain the appearance of the scene from unobserved viewpoints, \model{} incorporates object-centric view synthesis priors~\cite{poole2022dreamfusion}. Unlike previous work, we composed the deformable Gaussians and jointly optimized them in a common scene-centric coordinate frame to better model occlusions and geometrical relationships. Thus, we equip each object with a sequence of transformations that map its Gaussians from object-centric to scene-centric coordinates, enabling joint optimization in a unified space.

\paragraph{Initializing Canonical 3D Gaussians.}
We initialize 3D Gaussians for each object by applying rendering losses to object-centered crops from the first video frame. To hallucinate unseen views and complete object geometry, we use Score Distillation Sampling (SDS)~\cite{poole2022dreamfusion}, producing dense 3D Gaussian representations within object-centric viewing spheres.

For the static background, we unproject the RGBD information into a 3D point cloud and initialize a separate set of Gaussians. These background Gaussians have fixed 3D positions but remain trainable in other properties (e.g., color, opacity) during optimization using only RGB rendering supervision.

\paragraph{Transforming Between Object-Centric and Frame-Centric Coordinate Systems.}

To bridge object-centric and frame-centric spaces, we estimate a set of per-object, per-frame transformations. These allow us to integrate object-centric priors (e.g., SDS) with scene-centric rendering losses during the optimization process. These transformations are initialized using pre-trained mask trackers~\cite{sam2} and video depth estimators~\cite{depthCrafter}.

Specifically, for each object in frame $t$, we fit a 2D bounding box $B_t$ to its segmentation mask. We then compute an affine warp $W_t$ that maps $B_t$ to a canonical bounding box in a virtual object-centric view. Geometrically, $W_t$  can be interpreted as:
(1) a translation that centers the object within a virtual viewing sphere, and
(2) a scaling operation that normalizes its size.

To resolve 3D translation ambiguity in the depth direction, we leverage temporally consistent depth estimates~\cite{depthCrafter}. We randomly select one object $j$ as the reference and compute the relative depth scaling factor for another object $i$ at each frame as
$k_i = \frac{D_i}{D_j},$
where $D_i$ and $D_j$ are the median depth values for objects $i$ and $j$, respectively. Then, for frame $t$, we update the 3D position $\mathbf{p}^i_t$ and scale $\mathbf{s}^i_t$ of the Gaussians for object $i$ using:
$\mathbf{p}^i_t = \mathcal{C}^r - \left( \mathcal{C}^r - \mathbf{\mu}^i_t \right) \times k_i, \quad \mathbf{s}^i_t = \mathbf{s}^i_t \times k_i,$
where $\mathbf{\mu}^i_t$ is the original 3D position, $\mathbf{s}^i_t$ is the scale, and $\mathcal{C}^r$ denotes the camera position. This depth-aware scaling ensures consistent object placement across varying depths in the 3D scene.

\paragraph{Multi-Object Joint Gaussian Splatting.}
\label{sec:joint_splat}

We model the deformation of each object-centric Gaussian using K-plane-based deformation networks that predict changes in 3D position, rotation, and scale. These predicted deformations are then used to jointly render and optimize the deformable 3D Gaussians across all objects. To efficiently capture the complex yet structured motion patterns across different object parts, which are often highly correlated and lie in a lower-dimensional subspace, we adopt the concept of motion bases~\cite{wang2024shape}. Specifically, the displacement of each Gaussian is expressed as a linear combination of a shared set of motion basis vectors, enabling compact representation and improved generalization. To preserve local structural rigidity during deformation, we further regularize changes in relative 3D distance and orientation between neighboring Gaussians, following \cite{luiten2023dynamic}. Finally, we transform the deformed Gaussians from a virtual frame-centric coordinate system to the camera coordinate system using the estimated camera pose at each timestep $t$, and render the scene accordingly. This allows us to disentangle the object motion from the camera motion.

To account for temporal lighting and color variation common in long videos, we allow Gaussian appearance to evolve over time by introducing an additional MLP head on top of the K-plane deformation representation~\cite{cao2023hexplane} to predict time-varying adjustments to the spherical harmonic (SH) coefficients. The spatial and temporal continuity induced by the K-plane structure encourages nearby Gaussians to share similar color dynamics, providing a strong inductive bias. To avoid overfitting and prevent the network from explaining all appearance changes through color alone, we apply an L1 regularization term $\mathcal{L}_\text{reg}$ on the SH adjustments. 

A combination of losses is minimized to optimize the deformation network, including RGB rendering loss $\mathcal{L}_\text{rgb}$~\cite{kerbl20233d}, optical flow rendering loss $\mathcal{L}_\text{flow}$~\cite{dreamscene4d, wang2024shape}, depth consistency loss, and instance segmentation alignment loss, and the two regularization losses detailed above. For the instance segmentation alignment loss, each Gaussian is assigned a one-hot instance label as a fixed attribute, and we extend the differentiable 3D Gaussian renderer from~\cite{kerbl20233d} to render these instance labels, producing a segmentation map $\hat{y}_k$ for each frame. This rendered map is compared with the 2D instance masks $y_k$ from object trackers using a standard negative log-likelihood loss:
$\mathcal{L}_\text{class} = - \sum_k y_k \log{\hat{y}_k}$. This prevents Gaussians from drifting between objects during joint optimization.

\paragraph{Implementation Details.}

All experiments were conducted on a 48GB NVIDIA A6000 GPU. For object-centric lifting, we crop and scale the individual objects to approximately 65\% of the image size. Static 3D Gaussian optimization is performed over 1,000 iterations with a batch size of 16, consistent with prior work~\cite{ren2023dreamgaussian4d, dreamscene4d}. For deformation optimization, we run for 40× the number of frames, using a batch size of 8 and gradient accumulation over 2 steps. Additional implementation details, including loss-term weighting, are provided in the supplementary material.
\begin{figure}[!t]
    \centering
   \includegraphics[width=0.48\textwidth]{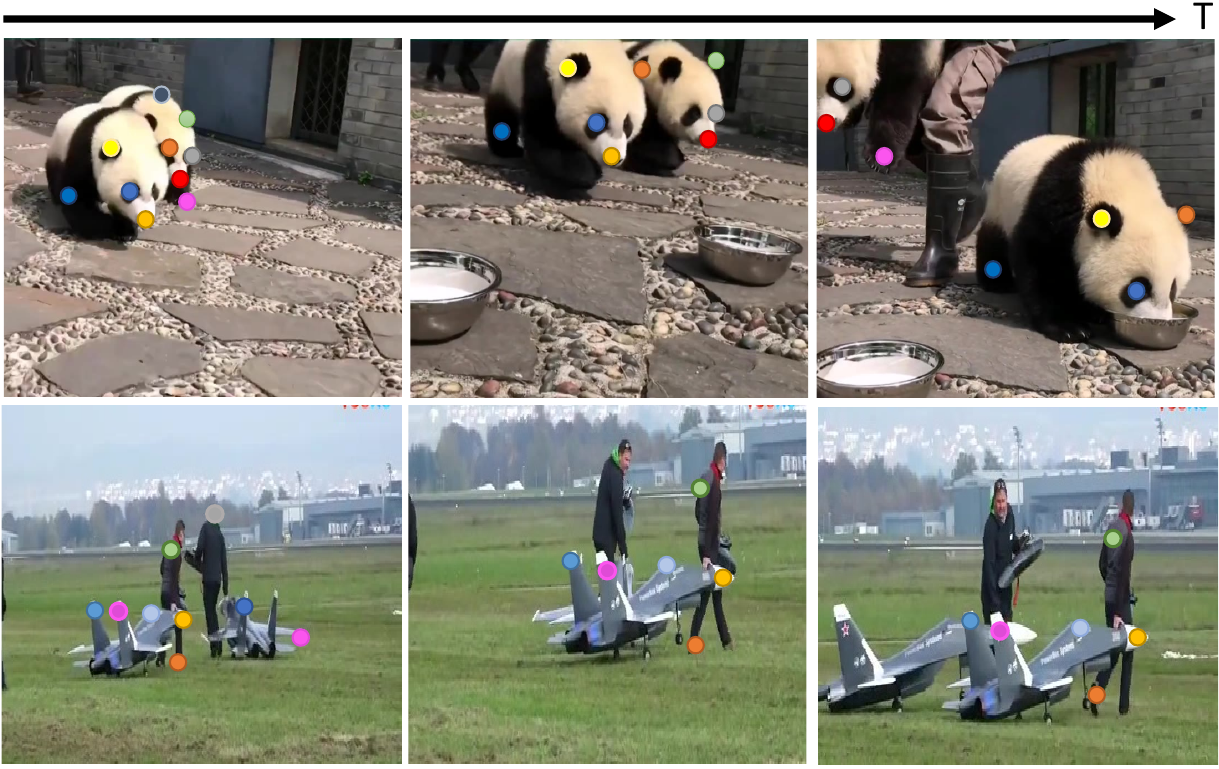}
    \caption{\textbf{Sample annotation from \dataset{}.} We visualize annotated points that are not occluded, with corresponding point tracks displayed in matching colors.}
    \label{fig:mose_anno_vis}
\end{figure}
\section{Experiments}
\begin{figure*}[!t]
    \centering
 \includegraphics[width=.99\textwidth]{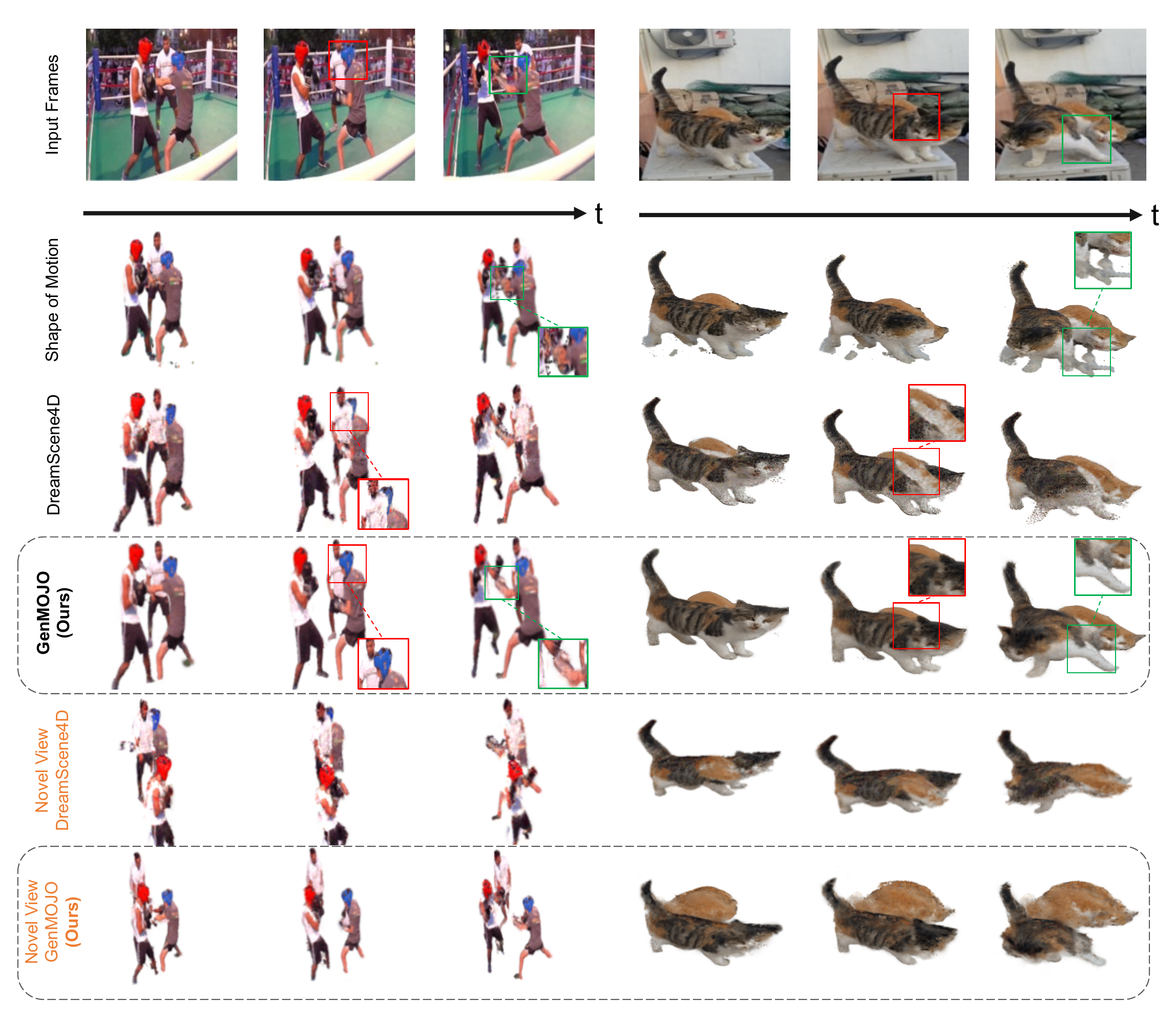}
    \vspace{-0.2in}
    \caption{{\bf Video to 4D Scene Generation Comparisons.} We render Shape of Motion~\cite{wang2024shape}, DreamScene4D~\cite{dreamscene4d}, and \model{} on a DAVIS video (left) and a MOSE video (right) from the reference view (top three rows) and a novel view (bottom two rows). We can see that the baselines produce artifacts or incorrectly model spatial relationships between objects, while our method does not exhibit such errors. We also render novel views for DreamScene4D and our method, where we can similarly observe improvements.}
    \label{fig:compare}
\end{figure*}
We test \model{} in 4D lifting from monocular videos. 
Our experiments aim to address these questions: 

\noindent \textbf{(1)} How does \model{} compare to other methods~\cite{dreamscene4d, jiang2023consistent4d, ren2023dreamgaussian4d} in terms of novel view synthesis? 

\noindent \textbf{(2)} How does \model{} perform in tracking points through occlusions compared to existing state-of-the-art supervised point trackers~\cite{karaev2023cotracker,karaev2024cotracker3} and test-time optimization~\cite{wang2024shape,dreamscene4d,ren2023dreamgaussian4d} methods? 

\noindent \textbf{(3)} To what extent do generative priors and object-based video decomposition contribute to improved view synthesis and point tracking?

\noindent \textbf{(4)} Do object-centric generative priors~\cite{liu2023zero} in \model{} outperform priors obtained from scene-level view-conditioned generative models~\cite{zeronvs,Rombach2022stablediffusion}?

To evaluate \model{} on complex, real-world scenarios, we select a challenging subset of 15 multiobject videos from DAVIS~\cite{pont20172017, doersch2022tap} and incorporate the 20 videos from our newly proposed \dataset{} dataset. This combined evaluation set spans a wide range of difficult scenes with diverse object motions and interactions.

\subsection{\dataset{} Dataset}
\label{sec:our_dataset} 
The proposed \dataset{} dataset includes 20 videos selected from MOSE~\cite{ding2023mose}, a dataset designed for complex video object segmentation with substantial real-world occlusions. We filter the videos using 2D visible object masks to exclude low-interaction scenes, ensuring meaningful object interactions and tracking challenges. Compared to datasets like TAP-Vid-DAVIS\cite{doersch2022tap}, which often feature salient and isolated objects, MOSE-PTS presents more complex, cluttered scenes. Annotators labeled up to five objects per video, with five key points per object, and all annotations were conducted at a high resolution of 1080p to ensure precision and detail. Some annotation samples are visualized in Figure~\ref{fig:mose_anno_vis}. Further annotation details can be found in the supplementary material.

\begin{figure*}[!t]
    \centering
    \includegraphics[width=0.78\textwidth]{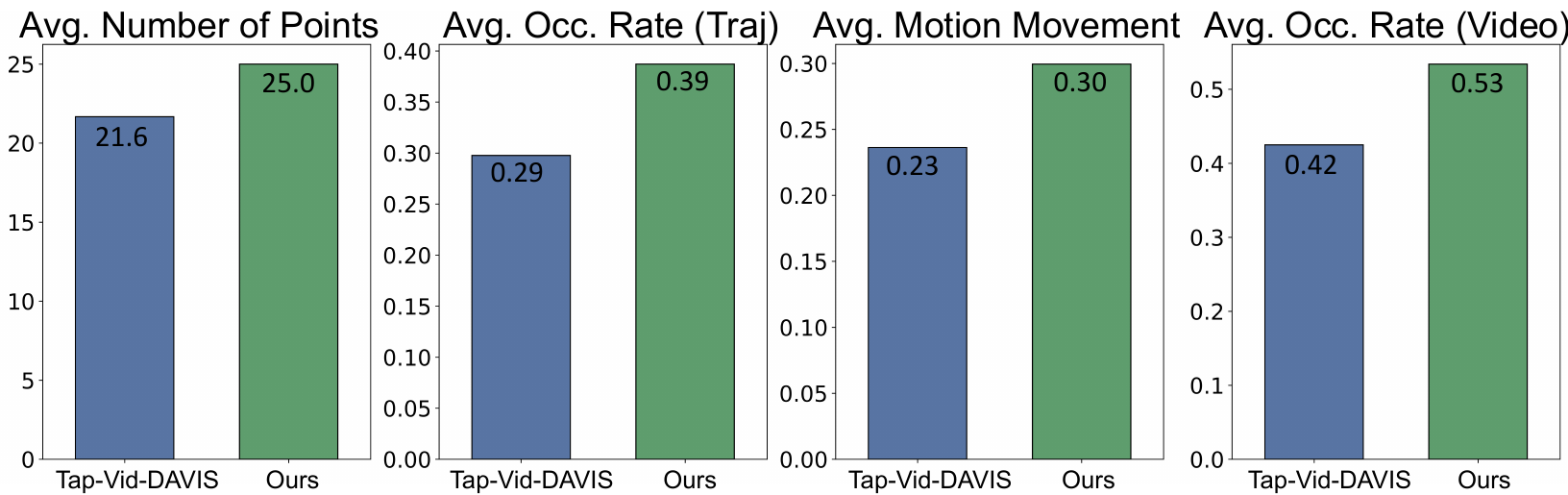}
    \caption{\textbf{Statistical comparison between \dataset{} and Tap-Vid-DAVIS~\cite{doersch2022tap}.} Avg. number of points: the labeled points in each dataset.  Avg. OCC. Rate (Traj / Video): the mean occlusion rate across individual trajectories, capturing the frequency of occlusions within distinct trajectories or videos. Avg. Motion Movement: the mean displacement of points between frames. These metrics show that \dataset{} contains more complex point tracks that have varied motion and undergo large occlusions.}
    \label{fig:stats}
\end{figure*}

\paragraph{Point Trajectory Analysis.}
In Figure~\ref{fig:stats}, we compare \dataset{} with TAP-Vid-DAVIS~\cite{doersch2022tap} based on four metrics: average number of points, occlusion rate per trajectory, motion range, and occlusion rate per video. \dataset{} has a higher average number of points and objects, which increases the complexity of the dataset. Additionally, \dataset{} has more complex multi-object occlusions and a wider motion range than TAP-Vid-DAVIS.

\subsection{Video to 4D Scene Generation}

\paragraph{Baselines.} We consider the following baselines for evaluating 4D scene generation: 

\noindent \textbf{(1)} Consistent4D~\cite{jiang2023consistent4d}, a method for video-to-4D generation from monocular videos that fits dynamic NeRFs per video using rendering losses and score distillation.

\noindent \textbf{(2)} DreamGaussian4D~\cite{ren2023dreamgaussian4d}, a 4D video generation method that, like our approach, uses dynamic Gaussian Splatting but does not employ object priors and thus lacks video decomposition.

\noindent \textbf{(3)} DreamScene4D~\cite{dreamscene4d}, a recent SOTA approach for 4D generation from multi-object monocular videos using dynamic Gaussian Splatting, making it the most directly comparable to our method. However, unlike \model{}, it optimizes each object’s motion independently, whereas \model{} jointly optimizes motion across all objects to capture object interactions.

\paragraph{Evaluation Metrics.}
The quality of 4D generation can be assessed in two key aspects: the view rendering quality of the generated 3D geometry and the accuracy of the 3D motion. This section focuses on view rendering quality. Following previous works~\cite{jiang2023consistent4d, ren2023dreamgaussian4d}, we report CLIP~\cite{radford2021learning} and LPIPS~\cite{zhang2018unreasonable} scores between four novel-view rendered frames and the input video frames, and calculate the average score per video. These metrics evaluate the semantic similarity between rendered and input video frames. Furthermore, we conducted a user study on MOSE videos, using a two-way voting method to compare \model{} with the state-of-the-art method DreamScene4D. Finally, we measure PSNR by comparing the rendered frame from the reference camera view to the input video sequence, providing a metric for the faithfulness of the 4D generation results.

\paragraph{4D Generation Results.} The 4D generation results on MOSE and DAVIS videos are presented in Table \ref{tab:exp_vis}. Since \model{} directly models objects in world space, it achieves the most faithful scene representation, as indicated by its high PSNR scores on both MOSE and DAVIS. \model{} also produces high-quality novel views, supported by its CLIP and LPIPS scores, which measure semantic similarity with the reference view, and by the user preference study. Consistent4D and DreamGaussian4D, originally designed for object-centric inputs, struggle with multi-object scenes like those in MOSE and DAVIS. This limitation often leads to distorted 3D geometry or artifacts due to ineffective motion optimization. While DreamScene4D can handle multi-object videos, it optimizes each object independently, which limits its ability to model inter-object occlusions or interactions in more complex videos.

We show qualitative comparisons on MOSE and DAVIS videos between in Figure~\ref{fig:compare}, highlighting instances where DreamScene4D fails to capture accurate relationships between interacting objects, leading to clipping problems. Additionally, we display reference view renderings from Shape of Motion, a recent Gaussian splatting-based method for point tracking. Since Shape of Motion only models visible parts of the scene, we exclude its results for novel views, and observe noisy renderings even in the reference view. These results demonstrate the robustness of \model{} in handling complex real-world videos with frequent occlusions and object interactions.

\begin{table*}[!htbp]
  \caption{\textbf{Video to 4D Scene Generation Comparisons}. We report the PSNR score for the re-rendered video under the reference camera pose, and the CLIP and LPIPS scores in novel views for MOSE~\cite{greff2022kubric} and DAVIS~\cite{pont20172017}. For user preference, we report the preference between DreamScene4D and \model{} in a two-way voting. \model{} outperforms existing methods for both novel view synthesis and produces faithful re-renderings in the reference camera view.}
  \label{tab:exp_vis}
  \centering
  \resizebox{0.8\linewidth}{!}{
  \begin{tabular}{lccccccccc}
    \toprule
    \multirow{2}{*}{Method} & \multicolumn{4}{c}{MOSE} & \multicolumn{3}{c}{DAVIS}\\
    \cmidrule(lr){2-5} \cmidrule(lr){6-8}
    & CLIP $\uparrow$ & PSNR $\uparrow$ & LPIPS $\downarrow$ & User Pref. & CLIP  $\uparrow$ & PSNR $\uparrow$ & LPIPS $\downarrow$\\
    \midrule
    Consistent4D~\cite{jiang2023consistent4d} & 77.78 & 22.59 & 0.172 & - & 77.17 & 22.31 & \textbf{0.146} \\
    DreamGaussian4D~\cite{ren2023dreamgaussian4d} & 81.96 & 17.82 & 0.195 & - & 81.62 & 17.68 & 0.183  \\
    {DreamScene4D~\cite{dreamscene4d}} & 85.16 & 22.98 & 0.169 & 36.8\% & 84.13 & 21.73 & 0.163 \\
    \midrule
    \model{} (\textbf{Ours}) & \textbf{85.41} & \textbf{25.56} & \textbf{0.168} & \textbf{63.2}\% & \textbf{84.17} & \textbf{23.51} & 0.163 \\
    \bottomrule
\end{tabular}}
\end{table*}

\begin{table*}[!t]
  \caption{\textbf{Point tracking comparison between 4D generation / reconstruction methods.} We report the Average Trajectory Error (ATE), Median Trajectory Error (MTE)~\cite{zheng2023pointodyssey}, Average End Point Error (A-EPE), and Median End Point Error (M-EPE)~\cite{dreamscene4d} in MOSE~\cite{greff2022kubric} and DAVIS~\cite{pont20172017, doersch2022tap}. \model{} outperforms existing methods that were not trained on point tracks in MOSE, and achieves competitive performance when compared against learning-based methods.}
  \label{tab:exp_motion}
  \centering
  \resizebox{0.8\linewidth}{!}{
  \begin{tabu}{lcccccccc}
    \toprule
    \multirow{2}{*}{Method} & \multicolumn{4}{c}{\textbf{MOSE}} & \multicolumn{4}{c}{\textbf{DAVIS}} \\
    \cmidrule(lr){2-5} \cmidrule(lr){6-9}
    & ATE $\downarrow$ & MTE $\downarrow$ & A-EPE$\downarrow$ & M-EPE$\downarrow$ & ATE $\downarrow$ & MTE $\downarrow$ & A-EPE$\downarrow$ & M-EPE$\downarrow$ \\
    \midrule
    \multicolumn{9}{l}{\textit{(a) \textbf{Not trained on point tracking data}}} \\
    \midrule
    GMRW~\cite{shrivastava2024self} & 24.85 & 20.87 & 42.59 & 36.99 & 42.17 & 26.73 & 75.61 & 60.14  \\
    DreamGaussian4D~\cite{ren2023dreamgaussian4d} & 27.73 & 25.71 & 46.18 & 43.55 & 30.27 & 28.18 & 61.5 & 55.69  \\
    Shape of Motion~\cite{wang2024shape} & 18.39 & 10.40 & 28.25 & 18.30 & 8.91 & \textbf{3.93} & 18.94 & \textbf{6.08}  \\
    DreamScene4D~\cite{dreamscene4d} & 13.47 & 10.49 & 22.48 & 16.97 & 9.59 & 6.50 & 17.47 & 10.29  \\
    \model{} (\textbf{Ours}) & \textbf{9.91} & \textbf{8.47} & \textbf{15.46} & \textbf{11.11} & \textbf{8.16} & 6.32 & \textbf{14.14} & 9.36  \\
    \midrule
    \multicolumn{9}{l}{\textit{(b) \textbf{Trained on point tracking data}}} \\
    \midrule
    \rowfont{\color{gray}}
    CoTrackerv2~\cite{karaev2023cotracker} & 15.74 & 12.15 & 27.96 & 20.35 & 21.90 & 2.08 & 26.22 & 3.49  \\
    \rowfont{\color{gray}}
    CoTrackerv3~\cite{karaev2024cotracker3} & 11.54 & 8.33 & 22.16 & 16.11 & 21.24 & 1.74 & 24.0 & 3.11  \\
  \bottomrule
  \end{tabu}}
\end{table*}

\subsection{4D Gaussian Motion Accuracy}
\paragraph{Baselines.} For evaluating Gaussian motion accuracy, we compare \model{} with DreamGaussian4D~\cite{ren2023dreamgaussian4d} and DreamScene4D~\cite{dreamscene4d}, as in the previous section. Consistent4D~\cite{jiang2023consistent4d} is not included since extracting accurate point motion from implicit NeRF representations is highly non-trivial and beyond our scope. We also include Shape of Motion~\cite{wang2024shape}, a recent optimization-based point-tracking method that employs 4D Gaussian Splatting. However, unlike \model{}, Shape of Motion does not leverage object or generative priors in optimizing dynamic Gaussians.

\paragraph{Evaluation Metrics.}
To evaluate motion accuracy, we use both Average Trajectory Error (ATE) and Median Trajectory Error (MTE)~\cite{zheng2023pointodyssey}, which measure the L2 distance between the estimated tracks and the ground truth tracks in all timesteps. MTE is included because it is less sensitive to extreme failure cases, providing a robust assessment of typical tracking accuracy. 
We also report the Average End Point Error (A-EPE) and Median End Point Error (M-EPE)~\cite{harley2022particle,dreamscene4d}, which calculates the L2 distance specifically at the final timestep. This provides specific insights into the tracking accuracy over long time horizons. For consistency, the predicted point tracks of all the methods are normalized to a resolution of $256 \times 256$ during the evaluation~\cite{doersch2022tap,karaev2023cotracker}.

\paragraph{Point Tracking Results.}
Table~\ref{tab:exp_motion} presents the point tracking results on MOSE and DAVIS video datasets, including comparisons with SOTA supervised methods~\cite{karaev2023cotracker, karaev2024cotracker3} trained on point annotations serving as an upper bound for performance. Supervised trackers like CoTrackerV2 and CoTrackerV3 show significantly lower accuracy on MOSE compared to DAVIS, highlighting the greater challenge posed by MOSE videos due to frequent, prolonged occlusions and larger object and point displacements. \model{} achieves the lowest tracking error in all semi-supervised and optimization-based methods, excelling in each evaluation metric. It also performs competitively on the simpler DAVIS dataset.

Both DreamScene4D~\cite{dreamscene4d} and \model{} benefit from object priors that help the Gaussians stay anchored to the same object throughout the video, enhancing tracking accuracy over other baselines, where tracked points often drift and get lost on different objects. This drift is a common issue for GMRW~\cite{shrivastava2024self}, Shape-of-Motion~\cite{wang2024shape}, and CoTracker~\cite{karaev2023cotracker}, especially during occlusions. However, DreamScene4D, unlike \model{}, lacks cross-object interaction modeling, which hinders its performance during occlusions. Without joint motion optimization, DreamScene4D’s Gaussian motion tracks in separate objects struggle to account for observed interactions in the video, causing tracks to drift or flicker within objects during occlusions, ultimately reducing tracking accuracy.

\subsection{Ablations}

We conduct ablations to understand how different components of \model{} affect generation quality and motion accuracy, as summarized in Table~\ref{tab:exp_ablation} using MOSE videos.

\noindent \textbf{Joint vs. Independent Object Optimization.} 
We first remove joint splatting and instead optimize each object separately, similar to DreamScene4D. This significantly reduces both motion accuracy and PSNR, showing that joint splatting is essential for handling occlusions and object interactions.

\noindent \textbf{Effect of SDS for Unseen Views.}
Next, we test what happens if we remove SDS supervision during motion optimization. This leads to a major drop in novel view quality and also hurts motion accuracy. Without SDS, Gaussians in unobserved views become under-constrained and tend to drift.

\noindent \textbf{Scene-level vs. Object-level SDS.} 
We compare object-centric SDS (used in \model{}) to using scene-level SDS with a scene-level novel-view synthesis model, ZeroNVS~\cite{zeronvs}. The object-centric version performs better, confirming that object-specific view synthesis provides more accurate guidance during optimization. 

\noindent \textbf{Instance Mask Rendering Loss.}
Finally, we remove the instance mask loss. Motion accuracy decreases, although the novel view quality stays about the same. This happens because some Gaussians drift between objects, leading to less precise tracking without this constraint.

\noindent \textbf{Depth Model Robustness.}
We replace DepthCrafter with DepthAnything-V2, which produces less stable depth predictions across frames. The impact on performance is minor, suggesting that \model{} is relatively robust to depth estimation noise.

\subsection{Limitations}
Despite the progress demonstrated in this paper, \model{} has several limitations: \textbf{(1)} Similar to previous methods~\cite{ren2023dreamgaussian4d, dreamscene4d}, our approach relies on the generalization capabilities of the view-conditioned generative model, which struggles with videos featuring unconventional camera poses. \textbf{(2)} Although our joint motion optimization procedure can correctly determine the depth ordering of objects, jittery depth estimations can still cause objects to ``jump” along the depth axis. \textbf{(3)} Due to the reliance on test-time optimization, our method is not suitable for online applications. We hope to develop novel feedforward models that can enable both real-time inference and provide stronger priors for test-time optimization in future work.

\begin{table}[!t]
  \caption{\textbf{Ablation experiments}. We report motion accuracy and view synthesis metrics in MOSE~\cite{greff2022kubric} by removing different proposed components in \model{}. We additionally replace object-centric SDS with scene-level SDS~\cite{zeronvs}, and replace DepthCrafter with DepthAnything-V2~\cite{depth_anything_v2} for depth initialization.}
  \label{tab:exp_ablation}
  \centering
  \resizebox{1.0\linewidth}{!}{
  \begin{tabu}{lcccc}
    \toprule
    \multirow{2}{*}{Method} & \multicolumn{4}{c}{\textbf{MOSE}} \\
    \cmidrule(lr){2-5}
    & A-EPE $\downarrow$ & M-EPE $\downarrow$ & CLIP $\uparrow$ & PSNR $\uparrow$ \\
    \midrule
    Full Model & \textbf{15.46} & \textbf{11.11} & \textbf{85.41} & \textbf{25.56} \\
    \midrule
    \textbf{w/o} Joint Splatting & 23.48 & 16.97 & 85.18 & 22.88 \\
    \textbf{w/o} SDS & 19.44 & 14.50 & 81.44 & 25.30 \\
    \textbf{w.} Scene-Level SDS~\cite{zeronvs} & 20.03 & 15.49 & 83.05 & 25.17  \\
    \textbf{w/o} Instance Mask Rendering & 16.77 & 12.49 & 84.88 & 25.19 \\
    \textbf{w.} DepthAnything-V2~\cite{depth_anything_v2} & 17.32 & 12.84 & 85.29 & 25.33 \\
  \bottomrule
  \end{tabu}}
\end{table}

\section{Conclusion}
We introduced \model{}, a novel framework for 4D scene reconstruction from monocular videos that addresses the core challenge of recovering persistent object representations in complex, real-world scenarios with multiple moving objects and heavy occlusions.  Unlike prior work limited to single-object or lightly occluded scenes, \model{} supports accurate 4D reconstruction across space and time in highly cluttered environments. It achieves this through a compositional strategy: jointly rendering all objects for occlusion-aware supervision while independently leveraging diffusion-based priors for object completion from novel viewpoints. Differentiable affine transformations unify object- and frame-centric representations into a coherent generative framework. 
Through extensive experiments on challenging datasets such as DAVIS and MOSE, we demonstrate that \model{} significantly outperforms state-of-the-art approaches in rendering fidelity, point tracking accuracy, and perceptual realism. 

\section*{Acknowledgments} 
This material is based upon work supported by an NSF Career award, an AFOSR Grant FA9550-23-1-0257, and TRI research. Any opinions, findings, and conclusions or recommendations expressed in this material are those of the authors and do not necessarily reflect the views of the United States Army, the National Science Foundation, or the United States Air Force.

{
    \small
    \bibliographystyle{ieeenat_fullname}
    \bibliography{main}
}

\clearpage
\appendix
\clearpage
\setcounter{page}{1}
\maketitlesupplementary

\begin{figure*}[ht!]
    \centering
\includegraphics[width=1.0\linewidth]{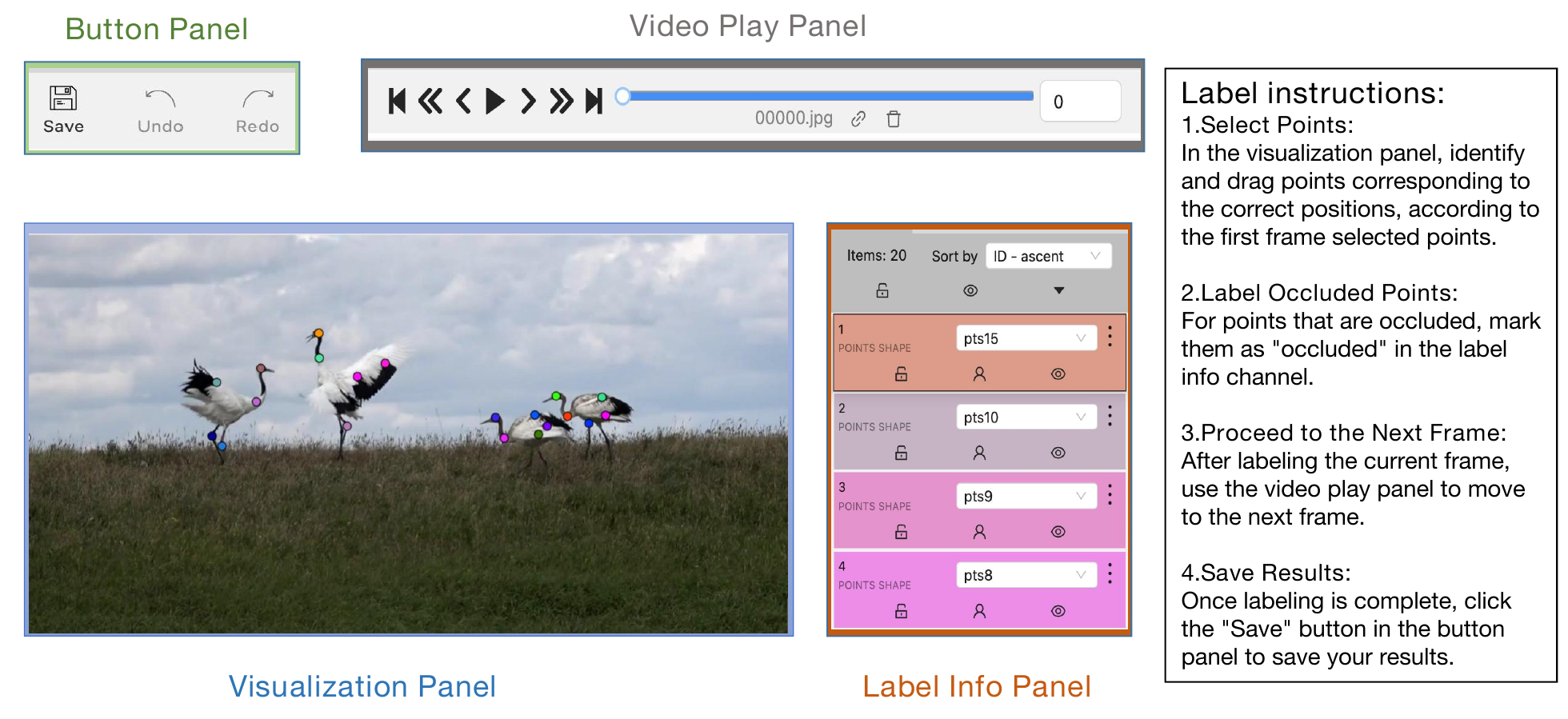}
    \caption{The annotation interface comprises four key components: a visualization panel for reviewing and marking points, button panel for interaction, an information panel displaying relevant details, and a video play panel for navigating through frames. The annotation process is structured into four iterative steps, guiding annotators to refine their annotations progressively.}
    \label{fig:supp_labeling}
\end{figure*}

\section{More Details on \dataset{}}

\subsection{Annotation Procedure}
We annotate 30 videos selected from MOSE~\cite{ding2023mose}, which is a complex and in-the-wild video dataset designed for video object segmentation with substantial real-world occlusions.
Inspired by the TAP dataset~\cite{doersch2022tap}, we aim for generality by allowing annotators to choose any object and any point they consider important, rather than specifying a closed-world set of points to annotate.

Given a video, annotation proceeds in two stages, depicted in Figure~\ref{fig:supp_labeling}.  First, annotators choose objects, especially moving ones, without regard to their difficulty in tracking.  Next, they choose points on each selected object and track them.  
Finally, we review and mark low-quality annotations for correction, repeating this correction procedure as many times as needed. All annotators provided informed consent before completing tasks and were reimbursed for their time.

\paragraph{Stage 1: Object Selection.}
For object selection, we began with the MOSE~\cite{ding2023mose} video mask annotations, initially identifying objects from these masks that predominantly captured moving elements within the scene. We then refined this selection by asking annotators to exclude overly simple objects, such as small, stationary, or occluded elements that contribute minimally to the scene’s complexity. To enhance the dataset’s challenge for point tracking, we included additional moving objects by human eyes, prioritizing those that experience partial occlusion during movement. This process ensured that the dataset maintained a higher level of complexity, better suited for evaluating robust point-tracking algorithms.

\paragraph{Stage 2: Point Annotation.}
For each object selected in the initial stage, annotators identified a set of five key points on the first frame, including prominent features such as eyes for animals and key poses for humans. To streamline the annotation process, we used CoTracker2~\cite{karaev2023cotracker} for point initialization across frames. Annotators then refined these points every alternate frame to ensure precise point tracking consistency with preceding frames, adjusting for minor shifts. In cases of occlusion, annotators marked frames where points were no longer visible, preserving accurate tracking continuity throughout the sequence.

\subsection{Point Track Difficulty Analysis}

To provide some insights into the difficulty of the annotated point tracks in \dataset{}, we run CoTrackerV2~\cite{karaev2023cotracker} and CoTrackerV3~\cite{karaev2024cotracker3} on \dataset{} and report the Occlusion Accuracy (OA; accuracy of occlusion prediction), Average Delta ($\delta$, fraction of visible points tracked within 1, 2, 4, 8, and 16 pixels, averaged over the 5 threshold values), and Average Jaccard (AJ, a combination of tracking accuracy and occludion prediction accuracy)~\cite{doersch2022tap}. For the evaluation on TAP-Vid subsets, we report the values reported in the original paper.

We find that in Table \ref{tab:supp_exp_motion} that both versions of CoTracker see up to 10 points of performance drop on \dataset{} across all metrics. Even the most recent CoTrackerV3, which has been trained on extra point annotations extracted from real world data, see a noticeable drop in performance on \dataset{}. This shows that \dataset{} is even more challenging than the videos presented in TAP-Vid-Kinetics~\cite{doersch2022tap}. We visualized our point track annotations in the \textbf{attached HTML page}, where we only show 15 video examples due to the Supplemental size limitation.

\begin{table}[!h]
  \caption{\textbf{Point tracking performance of CoTrackerV2 and CoTrackerV3 on various datasets.} The performance of CoTracker degrades by up to 10 points in \dataset{}, suggesting that it is a much harder dataset compared to the subsets present in TAP-Vid~\cite{doersch2022tap}.}
  \label{tab:supp_exp_motion}
  \centering
  \resizebox{1.0\linewidth}{!}{
  \begin{tabu}{l ccc ccc}
    \toprule
    \multirow{2}{*}{Dataset} & \multicolumn{3}{c}{\textbf{CoTrackerV2~\cite{karaev2023cotracker}}} & \multicolumn{3}{c}{\textbf{CoTrackerV3~\cite{karaev2024cotracker3}}} \\
    \cmidrule(lr){2-4} \cmidrule(lr){5-7}
    & AJ & $\delta$ & OA & AJ & $\delta$ & OA \\
    \midrule
    RGB-Stacking & 67.4 & 78.9 & 85.2 & 71.7 & 83.6 & 91.1  \\
    DAVIS & 61.8 & 76.1 & 88.3 & 63.8 & 76.3 & 90.2 \\
    Kinetics & 49.6 & 64.3 & 83.3 & 55.8 & 68.5 & 88.3 \\
    \midrule
    \dataset{} & \textbf{40.4} & \textbf{56.1} & \textbf{75.5} & \textbf{46.7} & \textbf{63.6} & \textbf{77.9} \\
  \bottomrule
  \end{tabu}}
  \vspace{-0.1in}
\end{table}

\section{Implementation Details}
\label{sec:rationale}

In this section, we provide some extra details on implementation and optimization, as well as the hyper-parameter choices for our experiments. We use the same set of hyperparameters for all DAVIS and MOSE videos in our experiments.

\subsection{Static Gaussian Optimization}
\paragraph{Pruning and Densification.}
Following \cite{tang2023dreamgaussian,ren2023dreamgaussian4d,dreamscene4d}, we prune Gaussians with an opacity smaller than $0.01$ or a scale larger than $0.05$. The densification is applied for Gaussians with an accumulated gradient larger than $0.5$ and max scaling smaller than $0.05$. Both pruning and densification are performed every $100$ optimization steps if the total number of Gaussians is smaller than $20000$. We impose this limit on the number of Gaussians due to memory constraints. Unlike traditional 3D Gaussian Splatting works, we do not reset the opacity values during optimization, which yields better results.
\paragraph{Hyperparameters.}
For static Gaussian optimization, we use the same set of hyperparameters from previous works~\cite{tang2023dreamgaussian, ren2023dreamgaussian4d, dreamscene4d} and use a learning rate that decays from $1e^{-3}$ to $2e^{-5}$ for the position, a static learning rate of $0.01$ for the spherical harmonics, $0.05$ for the opacity, and $5e^{-3}$ for the scale and rotation. We optimize for a fixed number of $1000$ steps for all objects in the video using a batch size of $16$ when calculating the SDS loss.
\paragraph{Running Time.} On the A6000 GPU, the static 3D-lifting process takes around 5.63 minutes for the $1000$ steps, comparable to what was reported in previous works.

\subsection{Dynamic Gaussian Optimization}

\paragraph{Deformation Network.} We follow previous works~\cite{ren2023dreamgaussian4d,dreamscene4d} and use a Hexplane~\cite{cao2023hexplane} backbone representation with a 2-layer MLP head on top to predict the required outputs. We set the resolution of the Hexplanes to $[64, 64, 64, 0.8T]$ for $(x, y, z, t)$ dimensions, where $T$ is the number of frames in the input video sequence. This is the same setting used in the baselines, and we keep this setting to ensure fair comparisons.
\paragraph{Hyperparameters.} The learning rate of the Hexplane is set to $6.4e^{-4}$, and the learning rate of the MLP prediction heads is set to $6.4e^{-3}$~\cite{ren2023dreamgaussian4d,dreamscene4d}. We optimize for a  $35 \cdot T$ steps with a batch size of $8$, where $T$ is the number of frames in the video. Like previous works, we sample $4$ novel views per frame in the batch to calculate the SDS loss. We use the AdamW optimizer to optimize the deformation parameters.

\paragraph{Running Time.} The running time is heavily dependent on the length of the video, as the number of optimization steps is dependent on the total number of frames, as well as the number of objects. For reference, on the A6000 GPU, a video with 32 frames with 3 objects takes around 74 minutes. DreamScene4D~\cite{dreamscene4d} is roughly 10\% slower as the optimization process needs to be done independently, but \model{} requires 15\% more VRAM on a GPU due to the joint splitting procedure requiring Gaussians of all objects to be kept in memory.

\section{Additional Results}

\begin{figure*}[!t]
    \centering
    \includegraphics[width=0.98\textwidth]{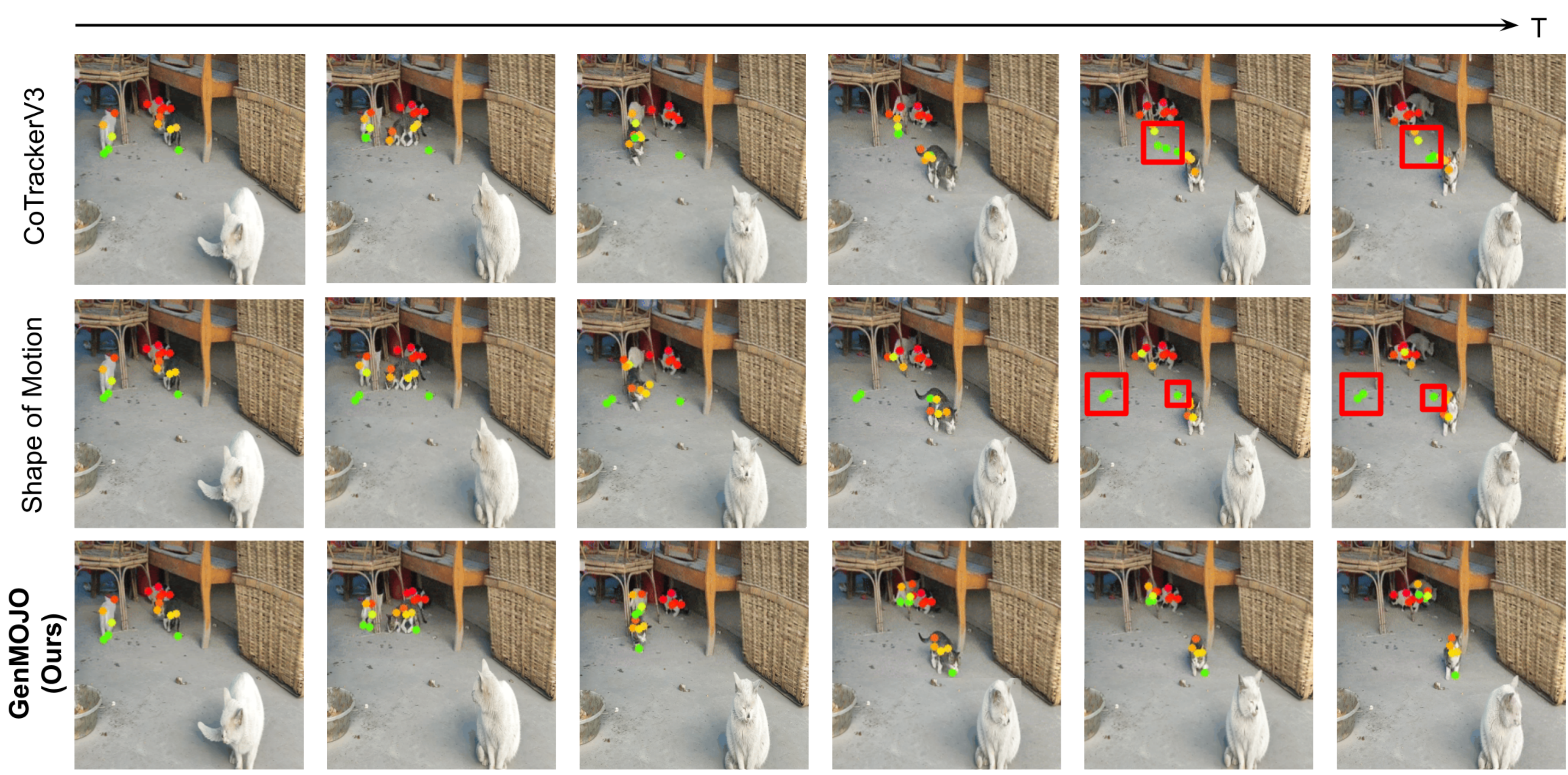}
    \caption{\textbf{Tracking comparisons.} \model{} produces more accurate point tracks compared to other optimization-based methods like Shape of Motion and DreamScene4D. CoTrackerV3 produces very accurate point tracks when it succeeds, but also produces tremendous errors when it fails, as the error can be unbounded.}
    \label{fig:tracking}
\end{figure*}

\subsection{Point Motion Visualizations}
We visualize some comparisons of the point tracking between Shape of Motion~\cite{wang2024shape}, \model{}, and CoTrackerV3~\cite{karaev2024cotracker3} in Figure~\ref{fig:tracking}. We can see the \model{} produces more accurate point tracks compared to other optimization-based methods like Shape of Motion and DreamScene4D. CoTrackerV3 produces very accurate point tracks when it succeeds, but also produces tremendous errors when it fails, as the error can be unbounded.

\begin{figure*}[!t]
    \centering
    \includegraphics[width=0.98\textwidth]{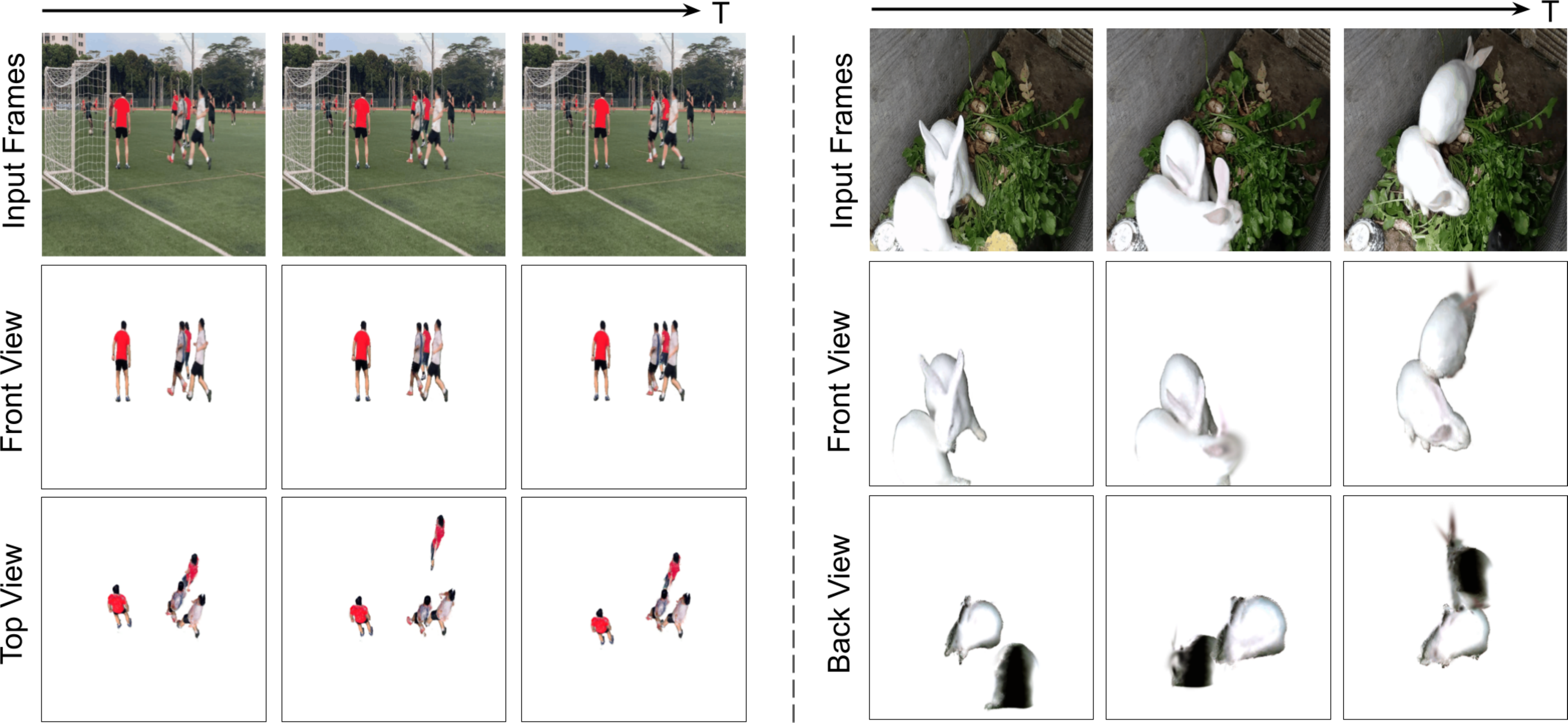}
    \caption{\textbf{Failure cases.} Top: Erroneous depth predictions can cause entities to jitter along the depth dimension. Bottom: View synthesis diffusion model failures result in degenerate textures in unseen viewpoints.}
    \label{fig:failure}
\end{figure*}

\subsection{Failure Cases}
We visualize some failure cases corresponding to the limitations section in the main text in Figure~\ref{fig:failure}. We envision that as better-performing foundational models for novel-view synthesis and video depth estimators are released, \model{} will also become more robust.

\section{Evaluation Details}
\subsection{Novel View Synthesis} We first recenter the Gaussians based on the median $(x, y, z)$, then render from the following combination of (elevation, azimuth) angles: $(0,\, 30)$, $(0,\, -30)$, $(30,\, 0)$, $(-30,\, 0)$, where $(0,\, 0)$ corresponds to the front view. These novel view renders are then compared with the reference view at each timestep to obtain the CLIP and LPIPS scores. We then average the scores to obtain the final score.

\subsection{User Study}
For the user study, we consider the videos in \dataset{}. For each method, we render from the reference view, as well as 2 novel views corresponding to the input video. We use Amazon Turk to outsource evaluations and ask the workers to compare the 2 sets of 3 videos with the input video. Each set of videos is reviewed by 30 workers with a HIT rate of over 95\% for a total of 900 answers collected. The whole user preference study takes about 89s per question and 22.25 working hours in total.

To filter out poor quality responses, we intentionally included ``dummy" questions where one set of renders was replaced by renders from non-converged optimization where the visual quality is noticeably worse, and used these ``dummy" questions to filter out workers who submitted responses that did not pass these questions. We additionally filtered out workers who submitted the same answer for all the videos. For every worker whose answers were rejected, we assigned new ones until the desired number of answers had been collected.

We include the full instructions shown to the workers below:

\begin{quote}
\textit{Please read the instructions and check the videos carefully.}
\bigskip

\textit{Please take a look at the reference video on the top. There are 2 sets of rendered videos (A and B) for this reference video: one rendered from the original viewpoint, and two rendered from other viewpoints. Please choose the set of videos that looks more realistic and better represents the original video to you. The options (A) and (B) correspond to the two sets of videos, as denoted in the captions under the videos.}

\textit{To judge the quality of the videos, consider the following points, listed in order of importance:}

\begin{enumerate}
    \item \textit{Do the objects in the rendered videos correspond to the reference video?}
    \item \textit{Do the objects clip or penetrate each other in the rendered videos?}
    \item \textit{Does the video look geometrically correct (e.g. not overly flat) when observed from other viewpoints?}
    \item \textit{Are there any visual artifacts (e.g. floaters, weird textures) in the rendered videos?}
\end{enumerate}

\textit{Please start from the first criteria to select the better set of renderings. If they are equal, please use the next criteria. If they are equal for all 4 points, please select Equally Preferred.}

\bigskip
\textit{\textbf{Please ignore the background in the original video.}}
\end{quote}

\begin{figure*}[!t]
    \centering
    \includegraphics[width=0.98\textwidth]{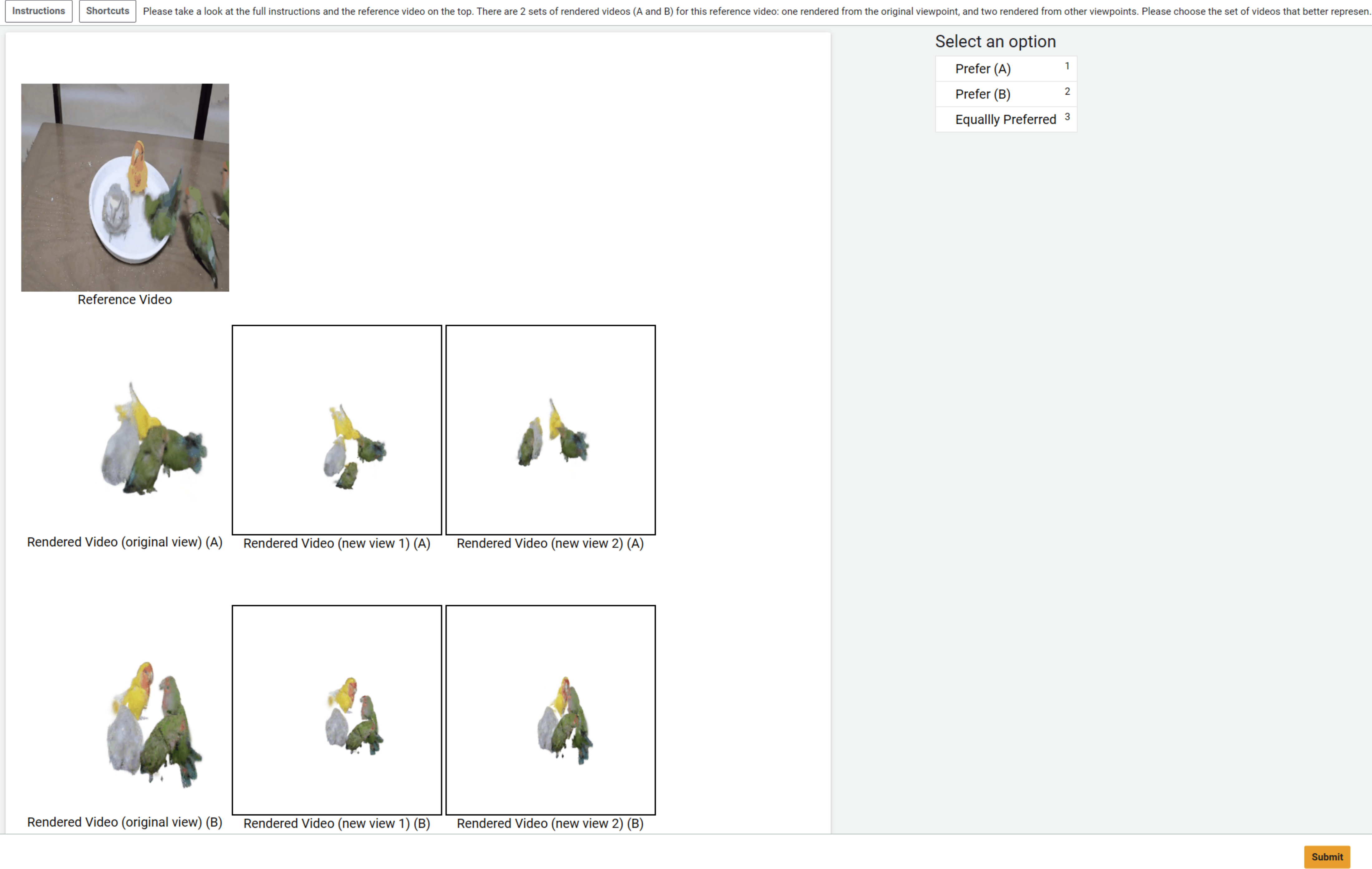}
    \caption{\textbf{User survey interface.} A GUI sample of what an Amazon Turk worker sees for the user study.}
    \label{fig:supp_survey}
\end{figure*}

A GUI sample of a survey question is also provided in Figure~\ref{fig:supp_survey} for reference.

\subsection{Multi-Object DAVIS Split}
We list the multi-object DAVIS video names that were used to perform evaluations:
\begin{verbatim*}
boxing-fisheye,car-shadow,crossing,
dog-gooses,dogs-jump,gold-fish, 
horsejump-high,india,libby,judo,
loading,pigs,schoolgirls,scooter-black,
soapbox
\end{verbatim*}

\end{document}